\documentclass[journal]{IEEEtran}
\usepackage{amsmath,amssymb,amsfonts}
\usepackage{algorithmic}
\usepackage{graphicx}
\usepackage{algorithm,algorithmic}
\usepackage{hyperref}
\usepackage{siunitx}
\hypersetup{hidelinks=true}
\usepackage{textcomp}
\usepackage{booktabs}
\usepackage{balance}
\usepackage{bm}
\usepackage{multirow}

\ifCLASSINFOpdf
\else
\fi
\begin{document}
\title{Refining Diffusion Models for Motion Synthesis with an Acceleration Loss to Generate Realistic IMU Data}

\author{Lars Ole Häusler$^{*}$,
        Lena Uhlenberg,
        Göran Köber,
        Diyora Salimova$^{\dagger}$,
        and~Oliver Amft$^{\dagger}$~\IEEEmembership{}% <-this % stops a space
\thanks{$^{*}$Corresponding author. $^\dagger$ Joint senior authors. L.O. Häusler, L. Uhlenberg, G. Köber are with the Intelligent Embedded Systems Lab, University of Freiburg, Germany, O. Amft is with Hahn-Schickard, Freiburg, Germany and the Intelligent Embedded Systems Lab, University of Freiburg, Germany (e-mail: haeusler@informatik.uni-freiburg.de, amft@ieee.org). D. Salimova is with the Department for Applied Mathematics, University of Freiburg, Germany (e-mail: diyora.salimova@mathematik.uni-freiburg.de).}}

\maketitle

\begin{abstract}
We propose a text-to-IMU~(inertial measurement unit) motion-synthesis framework to obtain realistic IMU data by fine-tuning a pretrained diffusion model with an acceleration-based second-order loss ($L_{\text{acc}}$). $L_{\text{acc}}$ enforces consistency in the discrete second-order temporal differences of the generated motion, thereby aligning the diffusion prior with IMU-specific acceleration patterns. We integrate $L_{\text{acc}}$ into the training objective of an existing diffusion model, finetune the model to obtain an IMU-specific motion prior, and evaluate the model with an existing text-to-IMU framework that comprises surface modelling and virtual sensor simulation. We analysed acceleration signal fidelity and differences between synthetic motion representation and actual IMU recordings. As a downstream application, we evaluated Human Activity Recognition~(HAR) and compared the classification performance using data of our method with the earlier diffusion model and two additional diffusion model baselines. When we augmented the earlier diffusion model objective with $L_{\text{acc}}$ and continued training, $L_{\text{acc}}$ decreased by 12.7\% relative to the original model. The improvements were considerably larger in high-dynamic activities~(i.e., running, jumping) compared to low-dynamic activities~(i.e., sitting, standing). In a low-dimensional embedding, the synthetic IMU data produced by our refined model shifts closer to the distribution of real IMU recordings. HAR classification trained exclusively on our refined synthetic IMU data improved performance by 8.7\% compared to the earlier diffusion model and by 7.6\% over the best-performing comparison diffusion model. We conclude that acceleration-aware diffusion refinement provides an effective approach to align motion generation and IMU synthesis and highlights how flexible deep learning pipelines are for specialising generic text-to-motion priors to sensor-specific tasks.

\end{abstract}


% \fi
\IEEEpeerreviewmaketitle

\section{Introduction}

\IEEEPARstart{I}nertial measurement units~(IMUs) are small, low-cost sensors that can measure at least linear acceleration and angular velocity, and are used in a variety of applications, including healthcare and clinical diagnosis~\cite{sarafSurveyDatasetsApplications2023, ricottiWearableFullbodyMotion2023}. A common task that uses IMU data is sensor-based HAR, which supports human behaviour understanding, safety and security monitoring, and daily routine tracking~\cite{gu_survey_2021, seiterDiscoveryActivityComposites2014}.
A main challenge of current HAR and IMU-based applications is the limited availability of supervised training data across the vast range of motion expressions. Since the collection, preparation, and annotation of IMU recordings require substantial expert involvement, datasets are constrained to specific settings and limited in the diversity of wearers. 

To address the common lack of training data, current research often incorporates synthetic data. In particular, cross-modality techniques have become popular for generating synthetic IMU measurements. Existing methods derive IMU data from video-based motion tracking~\cite{kwon_imutube_2020}, from biomechanical and surface models based on motion capture data~\cite{uhlenberg_synhar_2024}, or from motion generated using text descriptions~\cite{lengGeneratingVirtualOnbody2023, lengIMUGPTLanguageBasedCross2024,haeuslerText2IMUAdvancingHuman2025, haeusler_dynamic_2025}.
Compared to other approaches, text-based IMU synthesis is more versatile because it does not require collecting or processing new motion or sensor data for each application. For example, the recently introduced Text2IMU by Haeusler et al.~\cite{haeuslerText2IMUAdvancingHuman2025} used motion-synthesis models that generate motions directly from simple textual descriptions and synthesise IMU data, which are used solely for HAR model training to classify activities.

However, all current text-based IMU synthesis approaches treat the employed motion synthesis model as a fixed component. The motion synthesis models are primarily optimised for visual motion quality or diversity-related metrics and do not focus on the underlying acceleration of the generated motion, which should closely match that of real movement. Therefore, the motion prior itself should be adapted to IMU-oriented objectives such as providing accurate activity-dependent acceleration profiles, rather than a purely visual component.

In this work, we propose a framework that refines a pretrained diffusion model~\cite{tevet_human_2022} into an IMU-specific motion prior with an acceleration-based loss term. The refined model is then used to generate synthetic IMU data from textual activity descriptions. 

This paper provides the following contributions:
\begin{enumerate}
   \item We introduce an acceleration-based second-order term in the loss functions of a motion synthesis model. Since IMUs measure linear acceleration, the acceleration-based loss term is intended to explicitly align the generated motion with IMU signals.
   \item We finetune an existing motion synthesis model for IMU-specific motion generation and evaluate the resulting synthetic IMU data in a downstream human activity recognition task.

\end{enumerate}

\section{Related work}

\subsection{IMU data synthesis}
Synthetic data generation has become a widespread strategy to obtain sufficient training samples for complex HAR models when real IMU recordings are scarce or costly to acquire. Some research methods record or retrieve videos in which human motion is tracked and then translated into corresponding virtual sensors~\cite{huang_deep_2018, rey_let_2019}. 
Kwon et al.~\cite{kwon_approaching_2021, kwon_complex_2021, kwon_imutube_2020} proposed IMUTube, which translates online videos into IMU data by tracking individual joints using a 3D pose estimation model (VideoPose3D~\cite{pavllo3DHumanPose2019}). The results showed that fusing the synthesised data with real data considerably improved the recognition performance.

Instead of starting from videos, surface model simulations can be used to generate synthetic activity data. Compared to IMU datasets, large-scale collections of body surface motions~(e.g. AMASS~\cite{n_mahmood_amass_2019}) are available from MoCap technologies. Pei et al.~\cite{pei_mars_2021} used synthetic data generated by placing virtual sensors on the shape models from the AMASS dataset. Subsequently, the synthetic data were used to pretrain a deep learning model that was able to classify different motion categories substantially better than its non-pretrained counterpart. Uhlenberg et al.~\cite{uhlenberg_synhar_2024} utilised personalised biomechanical kinematic models and human surface models based on MoCap data to synthesise IMU sensor time series data, which improved HAR performance.
Recent advances in human motion synthesis propose cross-modality transfer as an alternative approach, i.e., mapping text descriptions and motion signals to generate synthetic IMU data~\cite{haeuslerText2IMUAdvancingHuman2025, lengGeneratingVirtualOnbody2023}.

In this work, we treat the text-to-motion setting as the core of our refined IMU data generation pipeline.

\subsection{Motion synthesis}

Human motion generation aims to generate natural human pose sequences that can be used in various real-world applications~\cite{zhu_human_2023}.
Increasingly popular is the text-to-motion task, in which natural language inputs are used to generate human motion in the form of joint rotations.
Large datasets, e.g., HumanML3D~\cite{guo_generating_2022}, include various motion sequences along with textual activity annotations. Using latent embedding of the textual input, models trained on such a dataset are able to create and combine a wide variety of human activities.
Multiple model architectures and concepts have been proposed to improve the human motion synthesis task, e.g., variational autoencoders that can be combined with a generative pre-trained transformer~\cite{zhang_t2m-gpt_2023}.
Another common approach are diffusion models, which use an iterative denoising process to generate motions. The Human Motion Diffusion Model (MDM)~\cite{tevet_human_2022} uses a transformer that directly predicts denoised motion samples. Further diffusion-based motion models include the masked modelling framework MoMask~\cite{guo_momask_2023} and ReMoDiffuse~\cite{zhang_remodiffuse_2023}, which introduces a retrieval module into the denoising process.

More recently, text-driven motion synthesis has been combined with IMU data generation. Leng et al.~\cite{lengGeneratingVirtualOnbody2023, lengIMUGPTLanguageBasedCross2024} proposed IMUGPT, a framework that employs the text-based human motion synthesis model T2M-GPT to generate synthetic accelerometer signals. Augmenting training with IMUGPT-generated data improves HAR performance compared to using only real IMU recordings. 
However, most synthetic IMU generation pipelines suffer from a domain gap between synthetic and measured sensor signals. Best performance is typically achieved only when synthesised data are calibrated or mixed with real IMU recordings from the target dataset~\cite{kwon_imutube_2020, zhang_unimts_2024, uhlenberg_synhar_2024, lengGeneratingVirtualOnbody2023}. 
Recently, Haeusler et al.~\cite{haeuslerText2IMUAdvancingHuman2025} proposed Text2IMU, a framework to generate synthetic IMU data from text and to train HAR models with a purely synthetic pipeline (Text Prompts $\rightarrow$ Motion Synthesis $\rightarrow$ Surface Model Generation $\rightarrow$ IMU Synthesis $\rightarrow$ HAR Prediction). Thus, data mixing or calibration of synthesised data with measured data of the target domain is no longer needed.

However, motion synthesis models are usually optimised for generic motion quality rather than IMU-specific characteristics, e.g., physically plausible linear accelerations. As a result, the generated motions may look realistic but still deviate from true IMU dynamics. 
Compared to previous work, our approach directly refines a pretrained diffusion-based motion prior with an acceleration-oriented loss that explicitly aligns the generated motion with the statistics of IMU measurements and thereby targets the domain gap at the motion prior level.

\section{Methods}
To generate synthetic data based on textual descriptions, we modified the framework proposed by Haeusler et al.~\cite{haeuslerText2IMUAdvancingHuman2025, haeusler_dynamic_2025}.
\subsection{IMU-tailored human motion synthesis}\label{sec:Human motion synthesis}
We used ChatGPT 4.1 to generate 10 different textual descriptions for each activity. We provided general structure, length constraints, and a few examples to include some variation in execution options similar to a real dataset. Execution variations depended on, e.g., speed and location of the person. For example, two generated prompts for walking are: "The person is walking on a track, following the oval path.", "The person is walking slowly with multiple strides.".

To synthesise human motion from the textual descriptions, we refined the Motion Diffusion Model~(MDM)~\cite{tevet_human_2022} to tailor it to IMU specific applications.
The MDM is based on stochastic diffusion processes that aim to synthesise a human motion $x^{1:N}= \{x^i\}_{i=1}^N$ of length $N$, represented either by joint positions or rotations, given a condition $c$.  Here, $x^i \in \mathbb{R}^{J \times D}$ for each $i \in \{1, \ldots, N\}$, where $J$ is the number of joints and $D$ is the dimension of the joint representation.  The model samples $x_0^{1:N}$ from the data distribution and for each noising step $t \in \{1, \ldots, T\}$ employs:
\begin{align}
\label{eq:diffuse}
    q(x_t^{1:N} | x_{t-1}^{1:N}) = \mathcal{N}(\sqrt{\alpha_t} x_{t-1}^{1:N}, (1-\alpha_t) I),
\end{align}
where $\alpha_t \in (0,1)$ is a constant parameter and $I$ is the identity matrix. From a theoretical point of view, the process corresponds to Euler discretisations of the underlying stochastic dynamics, and the discretisations themselves are considered as neural networks~\cite{GrohsJentzenSalimova2022}.
Given a condition $c$ random noise is sampled $x_T$ and iterates from $T$ to 1 at each step $t$ as follows:
\begin{enumerate}
    \item Transformer-encoder predicts the final clean motion $\hat{x}_0 = G(x_t, t, c)$,
    \item MDM diffuses is back to $x_{t-1}$ using \eqref{eq:diffuse}.
\end{enumerate}
The loss is defined to enforce physical properties and to encourage artefacts' natural and coherent motion as follows:
\begin{align}
    L_{\text{simple}} = \mathbb{E}_{x_0 \sim q(x_0|c), t \sim [1, T]} \bigl[ \|x_0 - G(x_t, t, c)\|_2^2 \bigr] 
\end{align}
and three additional common geometric losses that regulate  positions,  foot contact, and  velocities
\begin{align}
L_{\text{pos}} &= \frac{1}{N} \sum_{i=1}^{N} 
\bigl\| FK(x^i_{0}) - FK(\hat{x}^i_{0}) \bigr\|_2^2, \label{eq:lpos} \\
L_{\text{foot}} &= \frac{1}{N-1} \sum_{i=1}^{N-1} 
\bigl\| \bigl(FK(\hat{x}^{i+1}_{0}) - FK(\hat{x}^i_{0})\bigr) \cdot f_i \bigr\|_2^2, \label{eq:lfoot} \\
L_{\text{vel}} &= \frac{1}{N-1} \sum_{i=1}^{N-1} 
\bigl\| (x^{i+1}_{0} - x^i_{0}) - (\hat{x}^{i+1}_{0} - \hat{x}^i_{0}) \bigr\|_2^2. \label{eq:lvel}
\end{align}

For joint rotations, $FK(\cdot)$ denotes the forward kinematics function, which maps joint rotations to joint positions. In all other cases, $FK(\cdot)$ is the identity function. For each frame $i$, vector $f_i \in \{0,1\}^J$ represents the binary foot contact mask and indicates ground contact of the feet. The indicators are derived from binary ground-truth data~\cite{shiMotioNet3DHuman2020} and mitigate the foot-sliding effect by setting velocities to zero once the feet touch the ground.

To better align the motion prior with our IMU synthesis, we refined the loss function by adding an acceleration-based second-order term $L_{\text{acc}}$. The term directly penalises discrepancies in the discrete second-order temporal differences of the generated motion, which approximates joint accelerations, and is defined  as follows:
\begin{equation*}
    L_{\text{acc}} = \frac{1}{N-1} \sum_{i=2}^{N-1} \bigl\|(x_0^{i+1} - 2x_0^i + x_0^{i-1}) \\
    - (\hat{x}_0^{i+1} - 2\hat{x}_0^i + \hat{x}_0^{i-1})\bigr\|_2^2.
\end{equation*}

Here, the terms $(x_0^{i+1} - 2x_0^i + x_0^{i-1})$ and $(\hat{x}_0^{i+1} - 2\hat{x}_0^i + \hat{x}_0^{i-1})$ correspond to the second-order finite differences of the ground-truth and generated motion, respectively, and encourage the diffusion model to match the acceleration profiles that drive IMU signals. Overall, our training loss is defined as:

\begin{equation}
L = L_{\text{simple}} 
  + \lambda_{\text{pos}} L_{\text{pos}} 
  + \lambda_{\text{vel}} L_{\text{vel}} 
  + \lambda_{\text{foot}} L_{\text{foot}} 
  + \lambda_{\text{acc}} L_{\text{acc}}.
\label{eq:ltotal}
\end{equation}

As basis for fine-tuning our training loss, we used the MDM checkpoint provided by the authors that was trained with $T = 6000$ noising steps with DistilBERT text encoding.
We trained the model with our refined loss for an additional $T = 6000$ noising steps to obtain the final motion-synthesis model.
During fine-tuning, we kept the model architecture and training hyperparameters consistent with the original publication to examine the impact of our refined loss function. The final motion representation obtained from the textual descriptions consisted of a sequence of 22 joint positions in a 3D space with 20 samples per second, typically varying from 3-10 seconds.

\subsection{Surface model generation}\label{sec:Surface model generation}
To accurately simulate virtual sensor placements, the obtained motion representation%~(stick figure in Fig.~\ref{fig:teaser})
was converted into a surface mesh model. We use joints2smpl~\cite{zuo2021sparsefusion} to fit the 22 generated joints to a skinned vertex-based SMPL model~\cite{loper_smpl_2015}. The SMPL model consists of a triangulated mesh based on pose parameters $\theta$ and shape parameters $\beta$.

To simulate a diverse set of participants, we varied the SMPL shape parameter $\beta$ and generated multiple body models performing the same motions. We generated the shape parameters of 4 male and 4 female SMPL models using Virtual Caliper~\cite{pujades2019virtual}. Corresponding body height and body weights are shown in Tab.~\ref{tab:smpl_models}.

\begin{table}
    \centering
    \caption{Height and weight of eight SMPL models created from joints generated by T2M-GPT. SMPL parameters were chosen to represent a range of anthropometric sizes.}
    \begin{tabular}{c|c|c}
        %\hline
        \textbf{Model}       & \textbf{Height~(cm)} & \textbf{Weight~(kg)} \\ \hline
        Male tall & 200         & 100          \\ \hline
        Male short & 160         & 60          \\ \hline
        Male average & 178.9         & 85.8          \\ \hline
        Male adipose & 178.9         & 111          \\ \hline
        Female tall & 180       & 75          \\ \hline
        Female short & 150       & 50          \\ \hline
        Female average & 165.8        & 69.2           \\ \hline
        Female adipose & 165.8       & 97          %\\ \hline
    \end{tabular}
    \label{tab:smpl_models}
\end{table}

Participant's body characteristics were chosen to represent the tallest and shortest model for each gender~($|\beta_i| < 5, \quad \text{for all } i = 1, 2, \ldots, 10$), one model with average body measurements\footnote{\url{https://www.destatis.de/DE/Themen/Gesellschaft-Umwelt/Gesundheit/Gesundheitszustand-Relevantes-Verhalten/Tabellen/liste-koerpermasse.html}}, and one adipose model with average height but a high BMI of $35 \, \text{kg/m}^2$.

\subsection{IMU synthesis}\label{sec:IMU synthesis}
To synthesise IMU data and account for variations in sensor placement, nine vertices on the SMPL model surface mesh were selected, matching the IMU locations used in the RealWorld datasets. To derive synthetic acceleration and angular velocity sensor signals, we followed the approach described in Uhlenberg et al.~\cite{uhlenberg_comparison_2022,uhlenberg_synhar_2024}. 
Each selected vertex is described in a coordinate frame $G \in \mathbb{R}^{3 \times 3}$ by a position vector $\vec{k}(t) = (k_x(t), k_y(t), k_z(t))$ at time $t$. The linear acceleration at each vertex, $\vec{a}(t) = (a_x(t), a_y(t), a_z(t))$, is computed as the sum of the dynamic sensor acceleration $\vec{a}_d(t)$ and the gravitational acceleration $\vec{a}_g(t)$ acting on the sensor:
\begin{equation}
    \vec{a}(t) = \vec{a}_d(t) + \vec{a}_g(t).
\end{equation}

The dynamic acceleration is obtained as $\vec{a}_d(t) = d^{2}\vec{k}(t)/dt^{2}$, while the gravitational component is defined as $\vec{a}_g(t) = \vec{k}_j(t)\, g$ with $g = 9.81\,\text{m/s}^2$ and $\vec{k}_j(t)$ denoting a unit vector that represents the gravity along each sensor axis. To obtain $\vec{k}_j(t)$, the gravity-carrying axis $\vec{j}$ is transformed using the inverse of the rotation matrix $Q_{S}^{G}$, which represents the sensor orientation in frame $G$, i.e.\ $\vec{k}_j(t) = (Q_{S}^{G})^{-1} \vec{j}$. 

The angular velocity is synthesised from the time-varying quaternion orientation $q_{S}^{G}(t)$ associated with $Q_{S}^{G}$ according to:
\begin{equation}
    \vec{\omega}(t) = \frac{d\, q_{S}^{G}(t)}{dt}.
\end{equation}

\subsection{Evaluation}\label{sec:HAR evaluation}

To assess the impact of the introduced acceleration-based loss function, we first calculated the acceleration loss $L_{\text{acc}}$ for each training sample in the HumanML3D dataset before and after finetuning the model. The comparison quantifies the extent to which the refined motion prior has shifted toward IMU-consistent acceleration patterns.

Subsequently, we used the "\textit{all-MiniLM-L6v2}" pretrained model from SentenceTransformers~\cite{reimers-2019-sentence-bert} to generate embeddings of every sample text, which were compared to the eight activity categories common in both the HumanML3D~\cite{guo_generating_2022} and RealWorld~\cite{sztyler_-body_2016} dataset \textit{Dancing, Jumping, Lying, Running, Sitting, Stair climbing, Standing, Walking} or \textit{Other} using cosine similarity. The sample was assigned to the category with the highest similarity score. If the score did not exceed the threshold $\tau = 0.45$ for any of the categories, the sample was automatically assigned to \textit{Other}. We divided the categories into high-dynamic activities, characterised by larger acceleration/local maxima in signal amplitude and rapid orientation changes~(i.e., running, jumping, dancing), versus low-dynamic activities with smoother, lower-variance inertial patterns~(i.e., lying, walking, stair climbing, sitting, standing, and others).

We further evaluated our motion synthesis approach on the widely used public benchmark dataset, RealWorld~\cite{sztyler_-body_2016}.
We generated synthetic IMU data using simulated sensors positioned according to the placement descriptions provided in the datasets. Measured IMU data of the RealWorld dataset were downsampled to 20\,Hz to match synthetic data~(output of motion synthesis).

To examine the similarities and differences between the synthetic IMU datasets and the real IMU measurements from RealWorld, we applied dimensionality reduction for visualisation.
We computed 22 time-series features for each of the three axes of the first IMU~(Head) using the catch22 feature set~\cite{lubba_catch22_2019}.
We then used t-SNE~\cite{vanDerMaaten2008} to project all datapoints into a two-dimensional space to reveal structural patterns and potential discrepancies between real and synthetic IMU data.

As a downstream HAR application, we employed the DeepConvLSTM~\cite{ordonez_deep_2016} model architecture, which is widely used in HAR research~\cite{uhlenberg_synhar_2024, lengGeneratingVirtualOnbody2023, kwon_imutube_2020}. Input data was segmented into 2\,s windows~(40 samples per segment), using a sliding window with 50\% overlap.  Model training was carried out exclusively on the full synthetic dataset, whereas evaluation was conducted on the entire RealWorld dataset. All experiments were executed on a local computing cluster equipped with six NVIDIA A100 GPUs, and each experiment was repeated ten times.
For comparison, we provide IMU data generated from the original MDM checkpoint used in our finetuning. In addition, we provide HAR results from synthesised IMU data using the ReMoDiffuse~\cite{zhang_remodiffuse_2023} and MoMask~\cite{guo_momask_2023} diffusion models.

\section{Results}
\label{sec:results}
\begin{figure}[H]
     \centering
     \includegraphics[width=\linewidth]{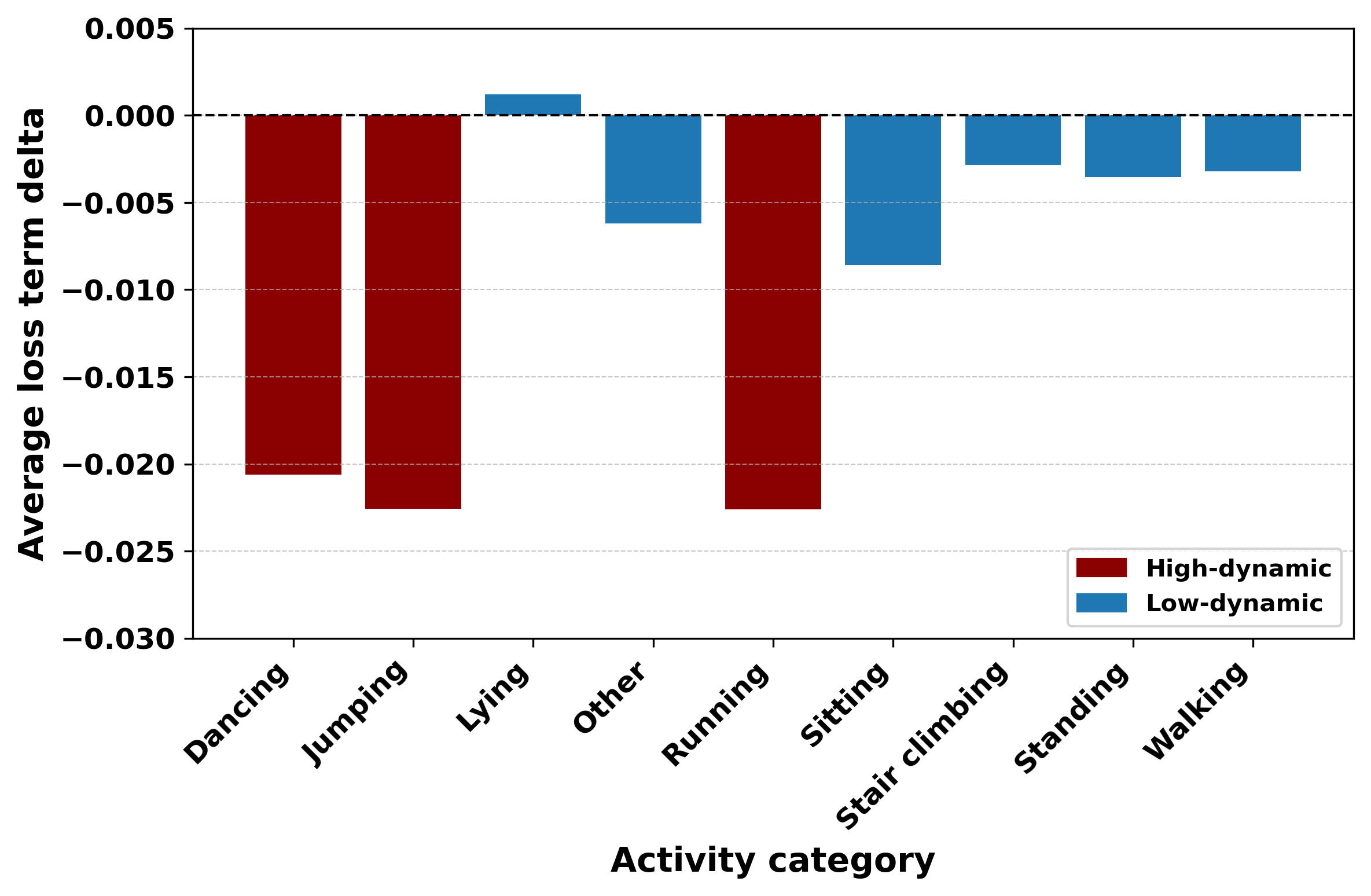}

     \caption{Average changes of the loss function when evaluating the MDM against our finetuned model on the HumanML3D dataset ($\Delta = \text{this work} - \text{MDM}$).}

     \label{fig:loss_delta}

 \end{figure}

Figure~\ref{fig:loss_delta} shows the average change of the acceleration loss $L_{\text{acc}}$ for different activity categories when comparing the earlier MDM with our finetuned model.
Overall, the results show that our approach achieved lower loss with $L_{\text{acc}}$ decreased by 12.7\% from the absolute value of 0.056 to around 0.049.
Since high-dynamic activities~(see Sec.~\ref{sec:HAR evaluation}) show more pronounced acceleration profiles, our tailoring leads to considerably larger reductions in $L_{\text{acc}}$. In contrast, low-dynamic activities, e.g., sitting, standing, and stair climbing, exhibited only small decreases in the loss, whereas for lying, the loss increased slightly.

\begin{figure}[H]
     \centering
     \includegraphics[width=\linewidth]{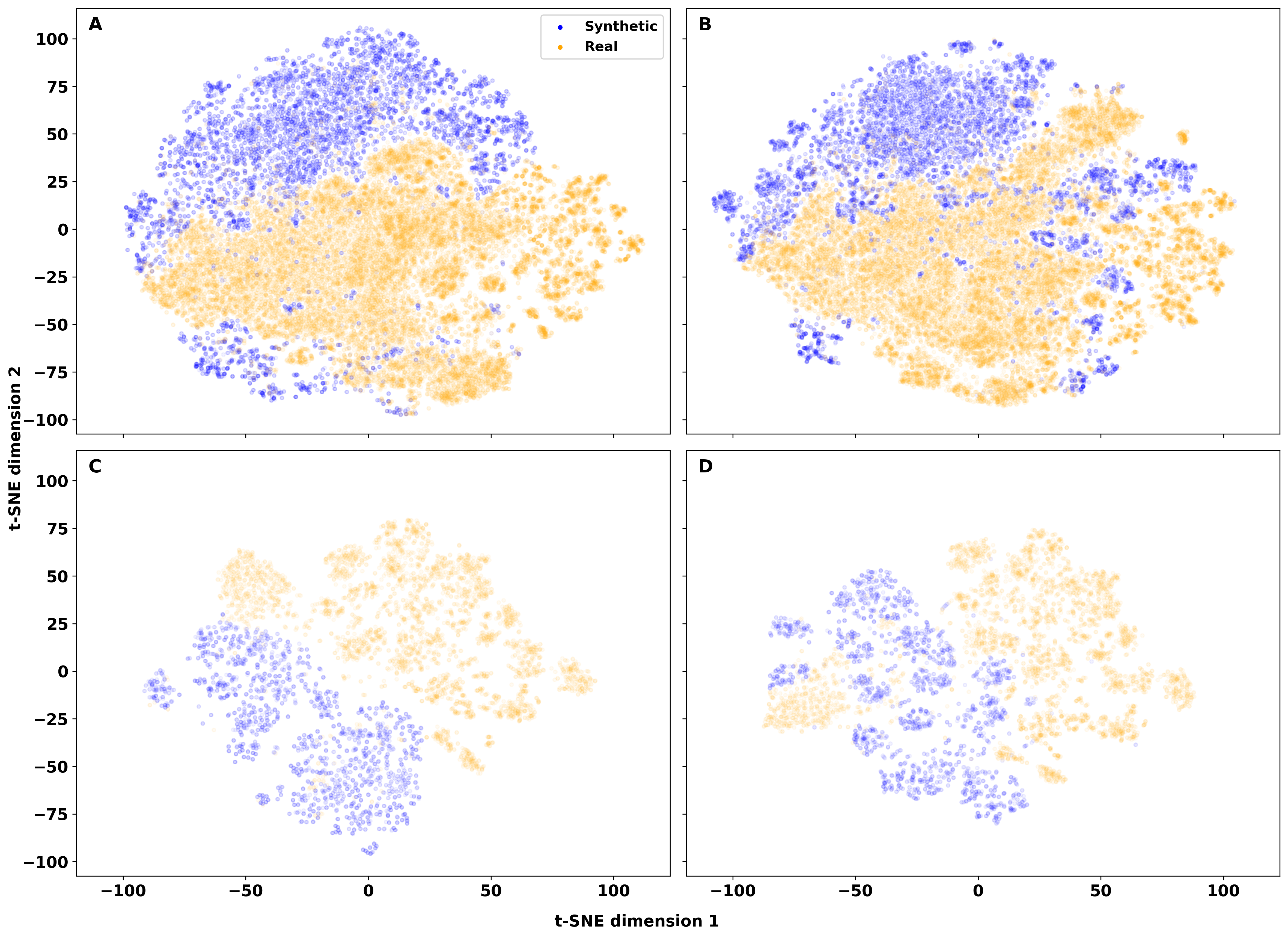}
      %vspace{-0.7cm}
     \caption{Comparison of dimensionality reduction results using t-SNE. A: MDM baseline using all activities B: Our proposed approach, including an acc term using all activities C: MDM baseline only jumping \& running D: Ours only jumping \& running.}
     \label{fig:tsne}
     %\vspace{-0.5cm}
 \end{figure}

Figure~\ref{fig:tsne} shows t-SNE embeddings of real and synthetic acceleration features for the MDM baseline and our proposed model. For all activities (Fig.~\ref{fig:tsne}A, B), our method achieved a higher overlap between synthetic and real data than the MDM baseline. 
When restricting the analysis to high-dynamic activities, i.e., jumping and running~(Fig.~\ref{fig:tsne}C, D), the effect of integrating the acceleration loss term into the training is more pronounced and a closer match between synthesised and real feature distributions is observed.

\begin{figure}[H]
     \centering
     \includegraphics[width=\linewidth]{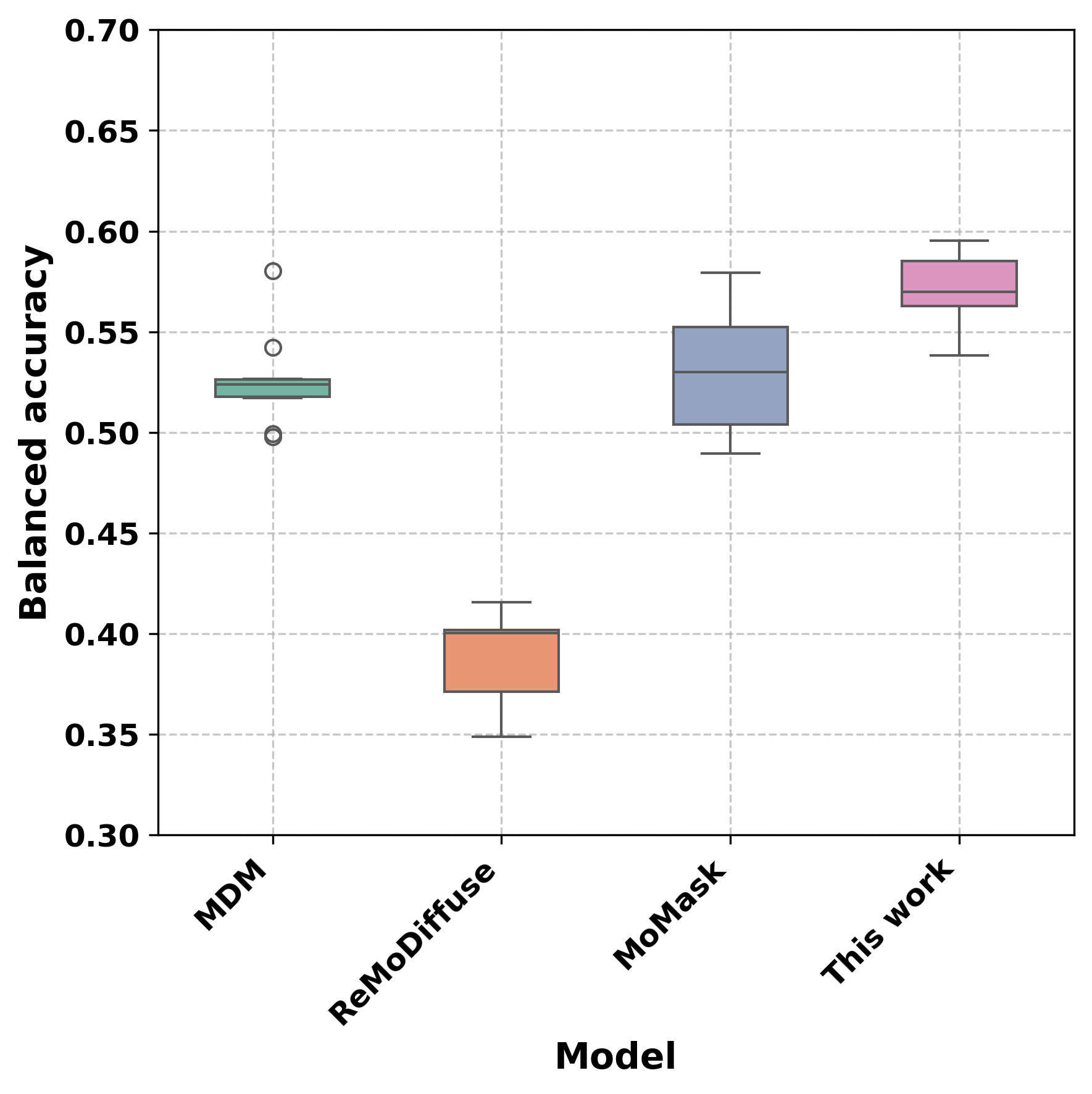}
     %\vspace{-0.7cm}
     \caption{Performance of DeepConvLSTM classification models on the RealWorld~\cite{sztyler_-body_2016} dataset. Models were trained exclusively with synthetic IMU data generated by different motion synthesis models, including the approach proposed in this work.}
     \label{fig:har_results}
     %\vspace{-0.5cm}
 \end{figure}

Figure~\ref{fig:har_results} shows the balanced accuracy for the HAR classifier trained with synthetic data from three baseline models compared to our method. The diffusion refinement, achieved by incorporating the acceleration-based second-order loss into the motion synthesis objective, resulted in higher HAR performance than all included comparison models~(52.6\% vs. 38.9\% vs. 53.1\% vs. 57.2\%). Thus, our approach yielded an increase of 8.7\% compared to the earlier MDM and 7.6\% compared to the best-performing model~(MoMask).

\section{Discussion}

Current end-to-end pipelines, including Text2IMU, increasingly comprise multiple components, and reliance on off-the-shelf pretrained modules inherently entails the risk of suboptimal performance. In this work, we showed that adapting pretrained pipeline components to the particular downstream sensor modality improves performance. In particular, we tailored a motion diffusion model from the computer vision domain by introducing an acceleration-based second-order loss term to match the requirements of our downstream applications based on IMU signals. The adaptation improved the realism of generated motions in the acceleration space and, consequently, increased HAR performance.

In more detail, our results showed that adding the second-order acceleration-based loss $L_{\text{acc}}$ aligned the motion synthesis prior with the acceleration profiles expected by IMU-based classifiers. Thus, by penalising acceleration inaccuracies in motion synthesis, the additional loss encourages the model to prioritise improvements in the realism of generated movements and subsequent IMU synthesis. The improvements were particularly pronounced for high-dynamic activities~(see Fig.~\ref{fig:loss_delta}), where large accelerations, pronounced amplitude maxima, and rapid orientation changes make accurate acceleration profiles critical for realistic IMU synthesis. The benefits of $L_{\text{acc}}$ were further evident in the IMU-data domain. The lower-dimensional representations of synthetic data aligned better with the empirical data, which indicates that the refined motion prior produces IMU features that are more consistent with those observed in real sensors, in particular, during high-dynamic activities~(see Fig.~\ref{fig:tsne}).  Results from the HAR classification experiment further supported the effectiveness of tailoring the motion synthesis model to the downstream task. Our fine-tuned model achieved higher balanced accuracy than the earlier MDM checkpoint and two other motion diffusion models~(see Fig.~\ref{fig:har_results}). However, the motion synthesis still did not accurately reproduce all activity classes, consistent with previous observations by Text2IMU~\cite{haeuslerText2IMUAdvancingHuman2025} and HAR performance remained below 60\%. Text2IMU~\cite{haeuslerText2IMUAdvancingHuman2025} reported that removing motions with low synthesis quality increased HAR balanced accuracy, which suggests that residual synthesis errors in the retained motions limited the achievable HAR performance.

Our motion diffusion refinement was intentionally conservative. We added a single acceleration-based term ($L_{\text{acc}}$), without systematic hyperparameter optimisation, architectural or scheduling modifications, or exploration of alternative weighting strategies for the individual loss components. Therefore, while severe overfitting to a particular dataset or HAR architecture is unlikely, our current results might underestimate the attainable performance increase. Thus, considerable potential remains for further improvements through more deliberate model and loss-function design, e.g., explore combinations of temporal smoothness, quality-aware losses, and task-driven losses, along with their weighting, and apply our refinement approach to additional motion synthesis models beyond MDM. 
In parallel to improved loss-function design, further improvements may be achieved by refining MDM sampling, which can be interpreted as Euler discretisations of the underlying stochastic differential equations. The analysis and stability of the discretisation error in conjunction with deep neural networks could provide a mathematical understanding of why and when diffusion models fail~\cite{GrohsJentzenSalimova2022}. Moreover, the errors could potentially be improved by using more advanced numerical schemes rather than merely Euler schemes~\cite{JentzenSalimovaWeltiBurgers}.

In addition, future work could move beyond local motion-level objectives that compare only motion synthesis model output with ground-truth trajectories and instead explicitly optimise the motion prior for end-to-end IMU synthesis and HAR accuracy by adding a HAR performance term to the fine-tuning loss.

Moreover, future work could address remaining synthesis-related limitations. For example, integrate more effective quality-control mechanisms to filter unreliable generated motions before HAR training or inference, as suggested by the observed performance increase when low-quality motions were excluded~\cite{haeuslerText2IMUAdvancingHuman2025}. Another option is to extend the training data with more diverse and representative recordings, particularly for classes that are currently synthesised inaccurately.

\section{Conclusion}
In this work, we refined a pretrained text-to-motion diffusion model to improve synthetic IMU generation for HAR. By introducing an acceleration-based second-order loss, we adapted the MDM from a primarily visual motion objective to one that more explicitly reflects the dynamics measured by wearable sensors. The refined model was integrated into our end-to-end Text2IMU pipeline.

Our evaluations show that the acceleration-aware adaptation improved the physical consistency of generated motions and narrowed the gap between synthetic and measured IMU characteristics. The effect is most apparent for high-dynamic activities, characterised by larger acceleration/local maxima in signal amplitudes and rapid orientation changes, for which accurate acceleration profiles are essential for realistic IMU synthesis. In addition, the improved alignment is reflected in feature-space analyses and translates into stronger generalisation in downstream HAR when training is performed solely on synthetic data. Compared with the earlier MDM model checkpoint and other diffusion baseline, our approach yields more reliable synthetic training data and better recognition performance on real-world test data.

Overall, our results indicate that acceleration-aware diffusion refinement represents an effective approach to align motion generation and IMU synthesis, and is a practical step towards robust, purely synthetic HAR pipelines. Our findings highlight the potential of specialising generic text-to-motion priors to sensor-specific tasks.

% or


%\section*{Acknowledgment}

\ifCLASSOPTIONcaptionsoff
  \newpage
\fi



%\bibliographystyle{IEEEtran}

\balance
\bibliographystyle{IEEEtran}
\bibliography{references}


%\vfill


\end{document}